\documentclass[10pt,twocolumn,letterpaper]{article}

\usepackage[pagenumbers]{wacv} 
\usepackage{wrapfig}
\usepackage{lipsum}  
\usepackage[most]{tcolorbox}
\usepackage{amsthm}
\usepackage{url}
\usepackage{multirow}
\usepackage{graphicx}   
\usepackage{booktabs}
\usepackage{tcolorbox}
\usepackage{csquotes}
\usepackage{algorithm}
\definecolor{wacvblue}{rgb}{0.21,0.49,0.74}
\usepackage[pagebackref,breaklinks,colorlinks,allcolors=wacvblue]{hyperref}

\def\confYear{2026}

\title{Unlocking Vision-Language Models for Video Anomaly Detection \\ via Fine-Grained Prompting}

\author{
    Shu Zou\textsuperscript{\rm 1} \qquad
    Xinyu Tian\textsuperscript{\rm 1} \qquad
    Lukas Wesemann\textsuperscript{\rm 2} \qquad
    Fabian Waschkowski\textsuperscript{\rm 2} \qquad\\
    Zhaoyuan Yang\textsuperscript{\rm 3} \qquad
    Jing Zhang\textsuperscript{\rm 1}
    \\
    \textsuperscript{\rm 1}Australian National University \quad \textsuperscript{\rm 2}Maincode \quad  \textsuperscript{\rm 3}GE Research
}

\begin{document}
\maketitle
\begin{abstract}
Prompting has emerged as a practical way to adapt frozen vision-language models (VLMs) for video anomaly detection (VAD). Yet, existing prompts are often overly abstract, overlooking the fine-grained human–object interactions or action semantics that define complex anomalies in surveillance videos. We propose \textsc{ASK-Hint}, a structured prompting framework that leverages action-centric knowledge to elicit more accurate and interpretable reasoning from frozen VLMs. Our approach organizes prompts into semantically coherent groups (\eg~violence, property crimes, public safety) and formulates fine-grained guiding questions that align model predictions with discriminative visual cues. Extensive experiments on UCF-Crime and XD-Violence show that \textsc{ASK-Hint} consistently improves AUC over prior baselines, achieving state-of-the-art performance compared to both fine-tuned and training-free methods. Beyond accuracy, our framework provides interpretable reasoning traces towards anomaly and demonstrates strong generalization across datasets and VLM backbones. These results highlight the critical role of prompt granularity and establish \textsc{ASK-Hint} as
a new training-free and generalizable solution for explainable video anomaly detection.
\end{abstract}

\section{Introduction}

\begin{figure}
    \centering
    \includegraphics[width=\linewidth]{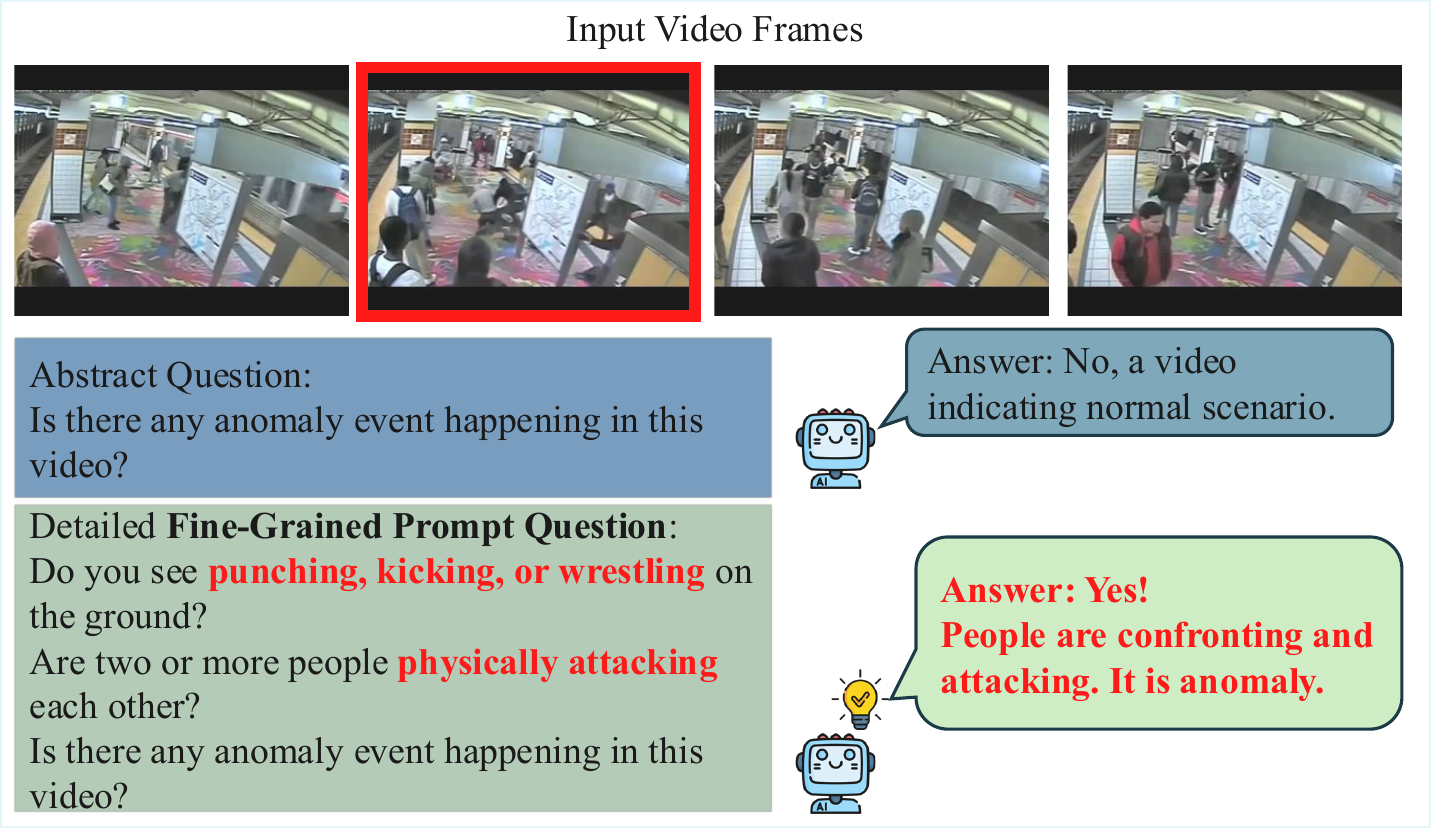}
    \caption{Performance of video anomaly detection \wrt
    prompt granularity.
    Given the same video input, an abstract prompt leads to a false prediction, while fine-grained action prompts (\eg~“punching”, “attacking”) elicit the correct abnormal classification from the model.
    }
    \label{fig:motivation:demo}
    
\end{figure}
Video anomaly detection (VAD) aims to automatically identify unexpected or abnormal events in video streams, which has found widespread applications in domains such as autonomous driving~\cite{bogdoll2023exploring} and surveillance monitoring~\cite{sultani2018real,zou2025simlabel}. Although improving detection performance is crucial, practical deployment often demands more than binary predictions (normal or abnormal). For instance, models must also provide interpretable reasoning behind their decisions, especially in high-stakes, open-world environments.
Recent advances in vision-language models (VLMs)~\cite{lin2023video, bai2025qwen2, wang2024internvideo2,tian2025thoughtaccuracydualnature} have shown great potential
in addressing these dual demands, showing its potential in downstream tasks~\cite{Tian_2024_CVPR,Tian_2025_CVPR,tian2025black}. By leveraging multi-modal architectures that combine powerful visual encoders with large-scale language reasoning capabilities, VLMs offer a new paradigm for VAD with natural language explanations.

To adapt VLMs for VAD, existing works can be broadly categorized into two streams. The first stream either decouples the process into visual captioning and external LLM-based reasoning~\cite{zanella2024harnessing,yang2024follow}, or fine-tunes VLMs via instruction tuning~\cite{lv2024video,zhang2024holmes} to jointly detect and explain anomalies in a black-box manner. 
While these methods demonstrate strong performance, they require significant computational cost, either at inference (due to external LLMs~/VLMs) or during training (due to full or partial model tuning).
The second stream focuses on adapting frozen VLMs by eliciting anomaly reasoning purely through prompt design. For example, VERA~\cite{ye2025vera} introduces a verbalized learning framework that learns a set of guiding questions from coarsely labeled data. 
However, its prompt optimization process operates in a black-box manner, where the guiding questions are updated via internal verbal feedback in implicitly performance-driven search while there is a lack of explicit control over their semantic structure or reasoning flow.
As a result, VERA~\cite{ye2025vera} offers limited interpretability and fails to fully exploit the compositional reasoning capabilities of VLMs.

A key observation is that humans rarely identify anomalies in videos based on abstract labels alone. 
When watching surveillance footage, we do not simply think ``this is robbery'' or ``this is an anomaly''; rather, we rely on perceiving fine-grained cues such as \emph{a person confronting another}, \emph{property being taken}, or \emph{an object being deliberately ignited}. 
These concrete human–object interactions allow us to rapidly and reliably recognize abnormal events. 
Similarly, for VLMs, abstract anomaly labels provide little visual grounding, whereas action-centric prompts offer explicit anchors that align language with visual evidence. 
Recent studies in video understanding highlight the importance of modeling fine-grained actions. TEAM~\cite{lee2025temporal} demonstrates that action-level matching improves few-shot recognition by using shared motion primitives. Video-R1~\cite{feng2025video} emphasizes step-wise reasoning over temporal segments for complex event understanding. 

Inspired by these findings, especially for abstract tasks such as detecting abnormal events, we raise a question: \textit{Can fine-grained prompting unlock stronger reasoning capabilities of VLMs for video anomaly detection?} (see Figure~\ref{fig:motivation:demo})
We thus conduct preliminary 
experiments over crime scenarios in the UCF-Crime dataset~\cite{sultani2018real} with VLMs-generated prompts~\cite{bai2025qwen2} to show how prompts of
different granularity affect
video anomaly detection. 
Particularly, given each anomaly class, we design three prompting strategies: 
\begin{itemize}
    \item \textbf{Coarse-Grained (Abstract):} 
    
    \texttt{Is there any anomaly event?}
    \item \textbf{Class-Label:} 
    
    \texttt{Is this video showing [Class Name]?}
    \item \textbf{Fine-Grained (Action-Centric):} 
    
    \texttt{Is there any fire or smoke? and etc.}
\end{itemize}

\begin{figure}
    \centering
    \includegraphics[width=\linewidth]{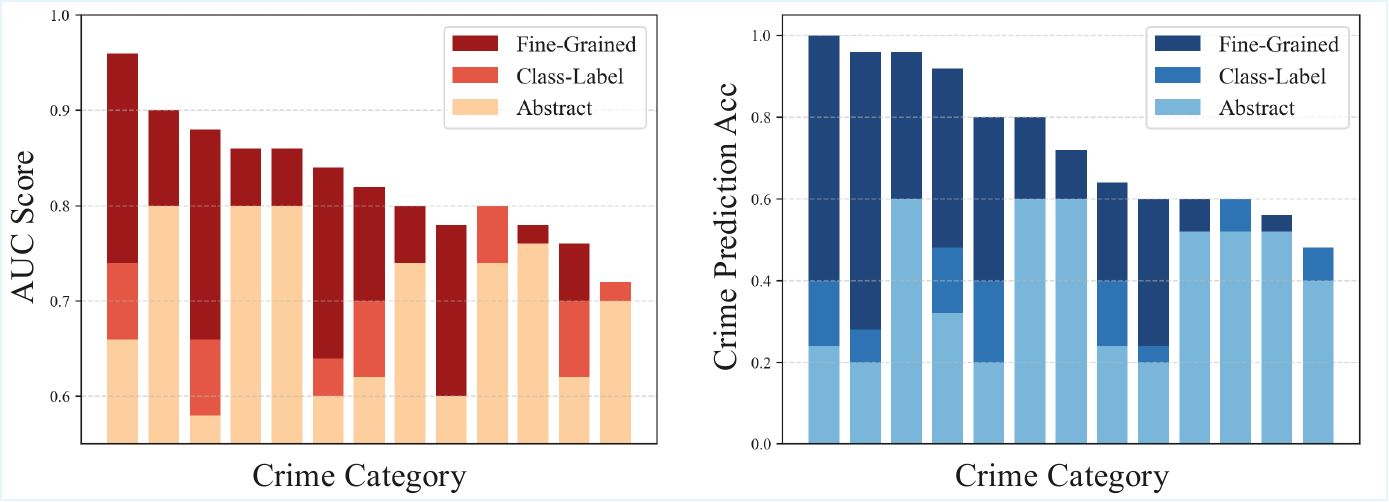}
    \caption{
        Comparison between coarse and fine-grained prompts across crime categories in the UCF-Crime dataset~\cite{sultani2018real}, where
        fine-grained prompts significantly improve AUC over corse prompts.}
    \label{fig:motivation_fine_vs_coarse}
    
\end{figure}

We evaluate (Figure~\ref{fig:motivation_fine_vs_coarse}) each strategy using a frozen VLM~\cite{bai2025qwen2}, measuring both the AUC score (left) and anomaly classification accuracy (right) across different crime categories.
We observe that fine-grained prompts and class-label prompts consistently outperform coarse prompts across nearly all crime categories. Remarkably, fine-grained prompting improves AUC by up to 30\% over abstract prompting, and leads to a substantial improvement in classification accuracy. These gains indicate that more fine-grained semantic information enables VLMs to better distinguish subtle or ambiguous abnormal behaviors while keeping high-performance on normal video prediction.

This motivates our design of a structured prompting framework that explicitly incorporates fine-grained action descriptions as a reasoning scaffold to explore the reasoning potential of VLMs for video anomaly detection. Our goal is to design an effective set of action-level prompts that generalize well while offering greater interpretability.
Taking a step further, inspired by~\cite{lee2025temporal}, we hypothesize and verify that some anomaly classes share underlying action patterns. For instance, “setting fire” is an intuitive cue for both “Explosion” and “Arson” (seeting details in Sec~\ref{sec:method}). Based on this intuition, we
leverage VLMs
to automatically analyze class-wise prompt sets and extract a compact set of \textit{shared, representative fine-grained prompts}. This compact set serves as the basis for adapting a frozen 
VLM to the VAD task, enhancing both efficiency and interpretability.



Our contributions are summarized as follows: \textbf{(1)}. We empirically show that fine-grained, action-centric prompts substantially enhance the reasoning capability of frozen VLMs for video anomaly detection (Figure~\ref{fig:motivation_fine_vs_coarse}). \textbf{(2)} We introduce \textbf{ASK-HINT}, a structured prompting framework that not only derives class-wise fine-grained prompts but also compresses them into a compact set of representative questions by exploiting shared semantic patterns across anomaly categories to unlock reasoning capabilities of VLMs with high-efficiency (Figure~\ref{fig:pipeline}). \textbf{(3)} We conduct extensive zero-shot evaluations on UCF-Crime~\cite{sultani2018real} and XD-Violence~\cite{wu2020not} datasets, demonstrating that ASK-HINT consistently surpasses prior baselines and establishes a new state-of-the-art training-free VAD solution with both stronger interpretability and robust generalization in cross-dataset and cross-class transfer settings (Section~\ref{Sec:experiment}). 

\section{Related Work}
\textbf{Video Anomaly Detection.}   
Existing solutions for VAD can be broadly categorized by the level of supervision.  
Supervised VAD~\cite{landi2019anomaly, liu2018future} requires frame-level annotations to train detection models, typically achieving high accuracy but incurring substantial labeling cost.  
Weakly-supervised approaches~\cite{wang2025federated, liu2024generalized, liang2025clip, lv2022spatio, mostafa2025abc} instead use video-level labels, offering lower annotation overhead but often lacking temporal precision or interpretability.  
Unsupervised methods~\cite{kobayshi2025unsupervised, liu2025multi, thakare2023rareanom, thakare2023dyannet, tur2023unsupervised} assume access only to normal videos and detect deviations from learned normality patterns, often relying on generative frameworks but struggling to generalize to diverse or unseen anomalies.
Recently, a new line of work explores open-world VAD using VLMs~\cite{zanella2024harnessing, yang2024follow, ye2025vera}. These approaches enable zero-shot inference and natural language explanation, opening up new possibilities for training-free and interpretable anomaly detection. Our work builds upon this direction, introducing a structured prompting framework that leverages fine-grained action cues to enhance reasoning and generalization.

\noindent\textbf{VLMs for Video Anomaly Detection.}  
VLMs have demonstrated strong capabilities in multimodal understanding and natural language reasoning. Recent studies have explored their application to VAD, leading to three major lines of work.  
The first integrates external large language models (LLMs) for enhanced reasoning~\cite{zanella2024harnessing, yang2024follow}, often formulating rule-based systems that combine visual captions with language-guided anomaly detection. While interpretable, these approaches require additional components and incur higher inference latency.  
The second line of work fine-tunes VLMs via instruction tuning or reinforcement learning to directly adapt them to anomaly detection tasks~\cite{zhang2024holmes, lv2024video, zhu2025vau}. These methods achieve strong performance but are resource-intensive, demanding substantial training data and computation.  
The third, and increasingly important, direction focuses on training-free VAD using frozen VLMs~\cite{shao2025eventvad, ye2025vera, lee2025flashback, zanella2024harnessing,cai2025hiprobe}. Among them, VERA~\cite{ye2025vera} introduces a verbalized learning framework that learns guiding prompts in a weakly supervised manner. However, its optimization process operates in a black-box fashion—lacking explicit semantic control, requiring external training, and offering limited interpretability.  
Our work builds on this training-free paradigm by introducing a structured prompting framework with fine-grained, action-centric prompts. This design leverages the compositional reasoning ability of VLMs, providing a potential research direction enabling both interpretability and effectiveness for adapting frozen VLMs to VAD tasks.

\section{ASK-HINT}\label{sec:method}

We propose ASK-HINT, a structured prompting framework for video anomaly detection using frozen VLMs. Built on the verified motivation that fine-grained, action-centric prompts yield more accurate and interpretable reasoning than coarse descriptions, ASK-HINT comprises three components: (1) class-wise prompt construction, (2) semantically prompt clustering and compression, and (3) structured inference with explanation trace. This design enables zero-shot and explainable VAD, while improving generalization to diverse and unseen anomaly types.

\begin{figure}[t]
    \centering
    \includegraphics[width=0.90\linewidth]{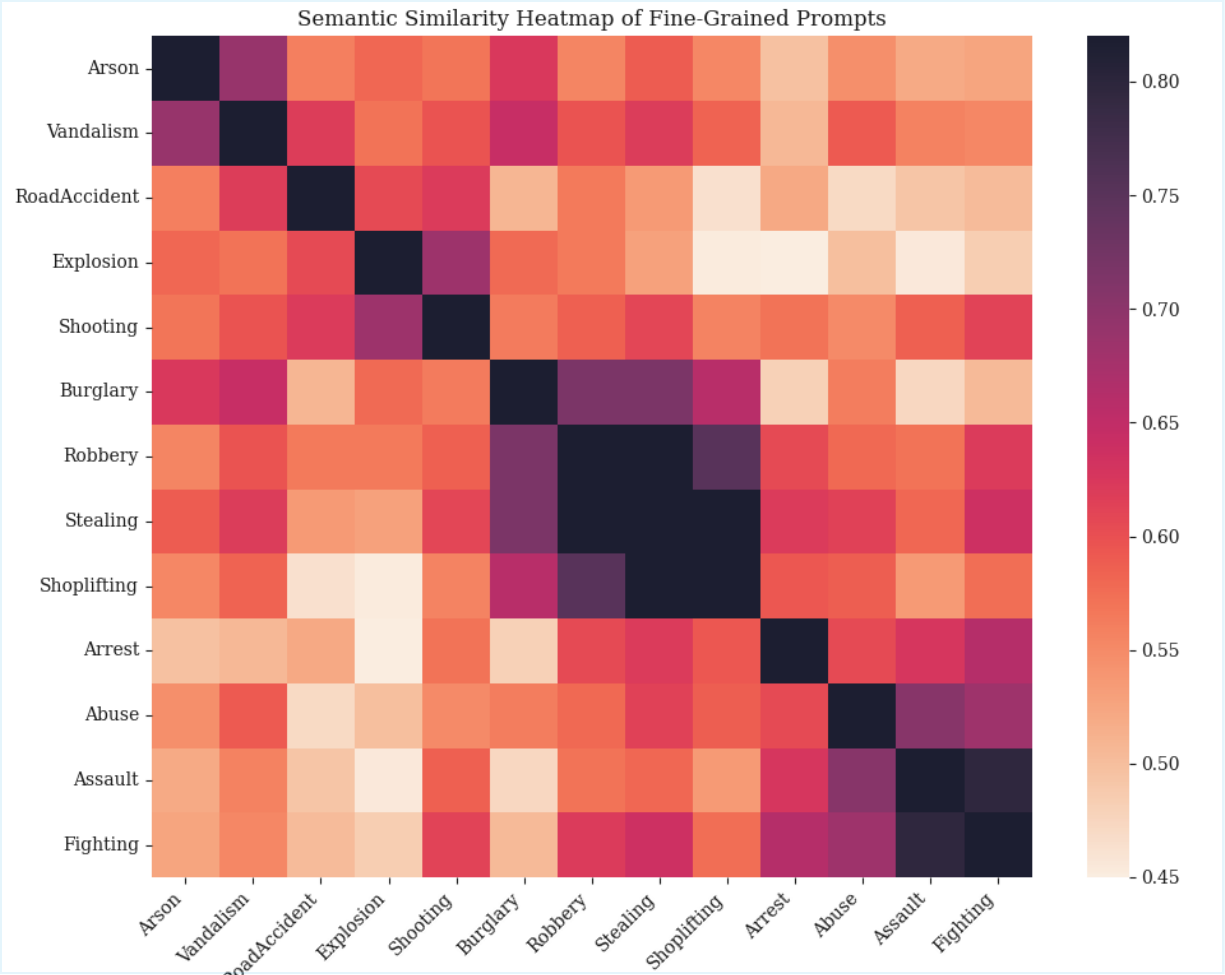}
    \caption{
        Prompt similarity heatmap across anomaly classes.  
        Cosine similarities between average prompt embeddings reveal semantically coherent clusters, such as \textit{Arson–Explosion} and \textit{Stealing–Shoplifting–Robbery}, supporting the hypothesis that anomaly categories share fine-grained action patterns.
    }
    \label{fig:heatmap-prompts}
    
\end{figure}

\subsection{Class-Wise Prompt Pool Construction}
\label{sec:prompt-pool}
We first construct a fine-grained prompt pool for each anomaly class, where
each prompt
is formulated as a natural language query that targets concrete visual actions or human–object interactions for the specific anomaly class. Prompts can be either manually designed or automatically generated and refined using LLMs (\eg~GPT-4~\cite{achiam2023gpt}) or VLMs (\eg~Qwen2.5-VL-7B-Instruct~\cite{bai2025qwen2}) based on class labels. In this paper, we leverage the strong capabilities of VLMs to construct a prompt pool for each anomaly class.


As a natural baseline, one may directly aggregate all the class-wise prompts and present the entire pool $\mathcal{Q}$
to the VLM during inference (See detailed prompt in Appendix~B). While this approach ensures maximal semantic coverage, using all class-wise fine-grained prompts during inference is inefficient and unsatisfactory (We report the performance of this baseline in Table~\ref{tab:full-prompt-comparison} as the \enquote{Full-Prompt Baseline}).
Existing work\cite{bae2025mash} explains that the poor performance may be caused by hallucination effects due to long prompts.
We therefore propose to further refine $\mathcal{Q}$ to reduce potential hallucination. In particular, we aim to identify a compact and generalizable subset of prompts
that capture the core action patterns across multiple anomaly categories.

\subsection{Semantic Compression via Prompt Selection}
\label{sec:semantic-compression}

Most existing prompt optimization approaches (\eg VERA~\cite{ye2025vera}) rely on performance-driven search, 
where candidate prompts are selected according to validation accuracy. 
While this strategy can be effective within a given dataset, it often yields prompts that are dataset- or video-specific, 
raising concerns about generalization to new scenarios. 
In contrast, we motivate our design from a semantic perspective: 
\textbf{ASK-HINT} derives prompts from fine-grained action semantics rather than validation scores, 
yielding a compact and transferable set of guiding cues. 
This encourages the model to focus on fundamental human–object interactions and recurring action primitives, 
which are more likely to generalize across unseen classes and datasets. 


\paragraph{Hypothesis: Shared Pattens Across Fine-Grained Actions.}
We hypothesize that many anomaly classes share underlying fine-grained action cues. For instance, ``setting fire'' is relevant to both \textit{Arson} and \textit{Explosion}, while ``physical confrontation'' frequently appears in \textit{Assault}, \textit{Robbery}, and \textit{Fighting}. This motivates compressing the full prompt pool $\mathcal{Q}$ into a smaller, representative set $\mathcal{Q}^*$ that retains discriminative power across related classes within the dataset.

\noindent\textbf{Empirical Validation.}
Given class-wise prompt pool, we encode them with the frozen Qwen2.5-VL text encoder, and compute pairwise cosine similarities between the average embeddings of different anomaly classes. As shown in Figure~\ref{fig:heatmap-prompts}, clear block-wise clusters emerge among semantically related categories (\eg, \textit{Arson–Explosion}, \textit{Robbery–Stealing–Shoplifting}). These clusters confirm the existence of shared semantic structures, providing a strong empirical foundation for prompt compression.

\begin{figure}[t]
    \centering
    \includegraphics[width=0.90\linewidth]{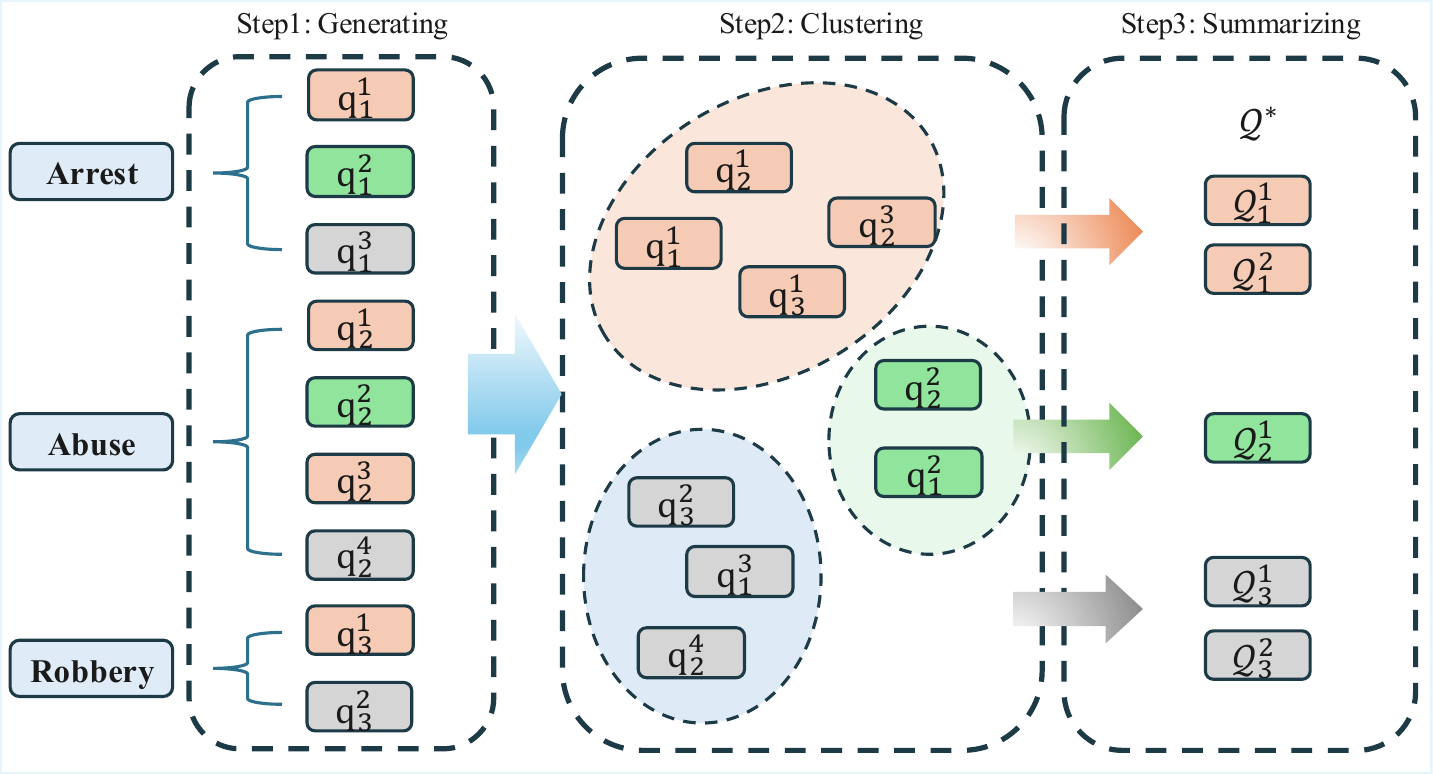}
    \caption{Overall pipeline of \textbf{ASK-HINT}. 
    Step~1: class-wise fine-grained action questions are generated for each anomaly category (\emph{Action Generation}).  
    Step~2: questions that reflect the same underlying action primitives are grouped together (\emph{Clustering}),which we mark with the same color.  
    Step~3: each cluster is condensed into representative guiding questions, yielding a compact and transferable prompt set $\mathcal{Q}^*$ (\emph{Summarizing}).  
   }
    \label{fig:pipeline}
\end{figure}

\noindent\textbf{Prompt Selection.}
Building upon these observations, we design a simple yet effective pipeline that leverages the semantic reasoning ability of VLMs to construct an \emph{optimal prompt set}. 
As illustrated in Figure~\ref{fig:pipeline}, the procedure consists of three steps:
\begin{enumerate}[label=\textbf{Step \arabic*:}]
    \item Generating the initial
    prompt pool $\mathcal{Q}$ according to Section~\ref{sec:prompt-pool}.
    \item Input $\mathcal{Q}$ into the VLM, which automatically reviews and clusters semantically related prompts into semantically related groups.
    \item For each group, summarize and generate 2--3 generalized guiding questions, forming the compact prompt set $\mathcal{Q}^*$.
\end{enumerate}

\begin{figure}[t]
    \centering
    \small
    \begin{tcolorbox}[colback=gray!5, colframe=black, title=ASK-HINT Prompt with $\mathcal{Q}^*$]
        \textbf{Task 1: Binary Decision.}  
        Using the questions in $\mathcal{Q}^*$ to classify a video as \textit{Normal} or \textit{Abnormal}.  

        \medskip
        \textbf{Task 2: Group Classification (if Abnormal).}  
        Based on the questions, assign the video to one of the following groups of questions:  

        \begin{itemize}
            \item \textbf{Violence or Harm to People}  
            \begin{itemize}
                \item Do you see people confronting, attacking, or restraining each other?  
                \item Is there evidence of weapons, force, or law enforcement?  
            \end{itemize}

            \item \textbf{Crimes Against Property}  
            \begin{itemize}
                \item Do you see someone unlawfully taking, concealing, or destroying property?  
                \item Do you see forced entry, vandalism, or deliberate fire?  
            \end{itemize}

            \item \textbf{Public Safety Incidents}  
            \begin{itemize}
                \item Do you see a sudden blast, smoke, or debris?  
                \item Do you see vehicles colliding or losing control?  
            \end{itemize}
        \end{itemize}

        \medskip
        \textbf{Answer Format:}  
        \begin{itemize}
            \item Normal Event. [short reason]  
            \item Abnormal Event $\rightarrow$ [Group]. [short reason]  
        \end{itemize}
    \end{tcolorbox}
    \caption{Video anomaly detection with the proposed
    \textbf{ASK-HINT}, using the UCF-Crime~\cite{sultani2018real} dataset as an example.
    It guides the VLM in two stages: (1) binary decision between normal and abnormal events, and 
    (2) group-level classification with justification.}
    \label{fig:prompt_template}
\end{figure}

We show our complete steps to generate optimal prompt set for UCF-Crime~\cite{sultani2018real} dataset in Figure~\ref{fig:prompt_template} and the detailed prompts are shown in Appendix C. Interestingly, the crimes are automatically grouped into three groups (Step 2): \enquote{Violence or Harm to People}, \enquote{Crimes Against Property}, and \enquote{Public Safety Incidents}. Given the intrinsic differences between the groups, we generate summarized questions for each group using a VLM (Step 3). The resulting $\mathcal{Q}^*$ highlights semantically broad or frequently recurring action patterns, enabling a more explicit reasoning trace toward \enquote{anomaly}. This VLM-guided compression reduces inference-time overhead, mitigates hallucinations from irrelevant prompts, and aligns prompt selection with the model’s intrinsic semantic understanding.


\subsection{Inference Procedure}
Given the optimal prompt set $\mathcal{Q}^*$, we design a structured prompting template (Figure~\ref{fig:prompt_template}) to guide frozen VLMs for inference in VAD. 
The core idea of \textbf{ASK-HINT} is to provide fine-grained action hints that help the model align language queries with visual evidence when deciding if a video contains abnormal behavior.  
During inference, the VLM makes decisions conditioned on the compact prompt set $\mathcal{Q}^*$ and generates structured outputs following the template. 
As illustrated in Figure~\ref{fig:prompt_template}, the model predicts whether the video is \textbf{Normal} or \textbf{Abnormal} (``Task~1''). 
If abnormal, it further assigns the video to one of the predefined semantic groups (\eg, \emph{Crimes Against Property}) and provides a concise rationale (``Task~2''). 
This design yields an interpretable anomaly detection pipeline, where both the decision and its justification are explicitly produced by the VLM.  
In contrast to validation-driven prompt optimization (\eg, VERA~\cite{ye2025vera}), \textbf{ASK-HINT} emphasizes semantic granularity and interpretability. 
By designing and compressing fine-grained prompts, our framework enables frozen VLMs to better exploit their reasoning ability without finetuning. 
This supports zero-shot evaluation across datasets and unseen categories, showing generalization potential without requiring extra training data or parameter updates.


%

\section{Experiments}\label{Sec:experiment}

We evaluate our proposed method, \textbf{ASK-HINT}, through a series of experiments designed to address two key questions: (1) How effective is it in eliciting the reasoning capabilities of frozen VLMs for video anomaly detection? (2) Given that method is designed as a general framework to enhance VLM performance on the VAD, how well does it generalize across different datasets and backbones?

\subsection{Experimental Setup}\label{sec:exper-setup}

\noindent\textbf{Datasets.} Following prior works~\cite{ye2025vera,zanella2024harnessing,shao2025eventvad}, we conduct evaluations on two standard VAD benchmarks:

\begin{itemize}
    \item \textit{UCF-Crime}~\cite{sultani2018real} is a large-scale surveillance video dataset containing 13 crime categories (e.g., Assault, Robbery, Arson) and one “Normal” category.
    
    \item \textit{XD-Violence}~\cite{wu2020not} is a multi-scene dataset with over 4,000 videos collected from movies, YouTube, and other online sources. The videos in the \textit{XD-Violence}~\cite{wu2020not} dataset may contain multi-labels, leading to 6 categories in total.
\end{itemize}

\noindent\textbf{Evaluation Metric.}  
Following prior work~\cite{zhang2024holmes,zanella2024harnessing,ye2025vera}, we adopt a  Area Under the Curve (AUC) score used to evaluate a model's ability in measuring the model's ability to distinguish between normal and abnormal events. Further more, we adopt accuracy (Acc) in measuring the ability of frozen VLM over crime detection.

\noindent\textbf{Comparison Methods.}  
We compare ASK-HINT against a broad range of existing methods, which we group into: (1) general VAD approaches and (2) VLM-based methods.

\begin{itemize}
    \item \textit{General Methods.}  
This includes weakly-supervised approaches~\cite{wu2020not,feng2021mist,tian2021weakly,wu2022self,li2022self,zhou2023dual,chen2023mgfn,wu2024open,joo2023clip,yang2024text,wu2024vadclip}, as well as self- and unsupervised methods~\cite{tur2023unsupervised,wang2019gods,zaheer2022generative,thakare2023dyannet}. These models typically rely on contrastive learning or reconstruction objectives trained on normal data.
\item \textit{VLM-Based Methods.}  
Recent methods explore VLMs for VAD, offering improved generalization and explanation capabilities. Some methods fine-tune the entire model or adapters (\eg~ VadCLIP~\cite{wu2024vadclip}, Holmes-VAU~\cite{zhang2024holmes}, HiProbe-VAD~\cite{cai2025hiprobe}), while others adopt a training-free setup, leveraging frozen backbones and natural language prompts (\eg~CLIP, LLAVA-1.5~\cite{liu2024improved}, VADor~\cite{lv2024video}, VERA~\cite{ye2025vera}, LAVAD~\cite{zanella2024harnessing}).
Our method, ASK-HINT, belongs to the training-free category and focuses on maximizing the reasoning capability of frozen VLMs via structured fine-grained prompts.
\end{itemize}

\begin{table}[t]
\centering
\small
\caption{AUC performance comparison on UCF-Crime
}
\renewcommand{\arraystretch}{1}
\renewcommand{\tabcolsep}{3.5mm}
\label{tab:ucf-crime}
\begin{tabular}{llc}
\toprule
\textbf{Training Type} & \textbf{Method} & \textbf{AUC\%} \\
\midrule

\multirow{11}{*}{\shortstack{Weakly \\ Supervised}} 
&XDVioDet~\cite{wu2020not}& 82.44  \\
&MIST~\cite{feng2021mist}& 82.30  \\
&RTFM~\cite{tian2021weakly} & 83.30  \\
&S3R~\cite{wu2022self} & 85.99  \\
&MSL~\cite{li2022self} & 85.62  \\
&UR-DMU~\cite{zhou2023dual} & 86.97  \\
&MFGN~\cite{chen2023mgfn} & 86.98  \\
&Wu et al.~\cite{wu2024open} & 86.40  \\
&CLIP-TSA~\cite{joo2023clip} & 87.58  \\
&Yang et al.~\cite{yang2024text} & 87.79  \\
&VadCLIP~\cite{wu2024vadclip} & 88.02  \\

\midrule
\multirow{3}{*}{Self Supervised
} 
&TUR et al.~\cite{tur2023unsupervised}& 66.85  \\
&BODS~\cite{wang2019gods}& 68.26  \\
&GODS~\cite{wang2019gods}& 70.46  \\
\midrule
\multirow{2}{*}{Unsupervised
} 
&GCL~\cite{zaheer2022generative}& 71.04  \\
&DYANNET~\cite{thakare2023dyannet}& 84.50  \\

\midrule
\multirow{2}{*}{\shortstack{Fine-Tuned\\ MLLM}} 
&Holmes-VAU~\cite{zhang2024holmes}& 87.68  \\
&HiProbe-VAD (Tuned)~\cite{cai2025hiprobe}& 88.91  \\
\midrule

\multirow{10}{*}{\shortstack{Training-Free\\MLLM}} 
&ZS CLIP~\cite{zanella2024harnessing}& 53.16  \\
&ZS IMAGEBIND-I~\cite{zanella2024harnessing}& 53.65  \\
&ZS IMAGEBIND-V~\cite{zanella2024harnessing}& 55.78 \\
&LAVAD~\cite{zanella2024harnessing}& 80.28 \\
&LLAVA-1.5~\cite{liu2024improved}& 72.84 \\
&VADor~\cite{lv2024video}&85.90 \\
&Holmes-vad~\cite{zhang2024holmes}&84.61 \\
&VERA~\cite{ye2025vera}&86.55 \\
&HiProbe-VAD~\cite{cai2025hiprobe}& 85.89\\
&\textbf{ASK-HINT(Ours)} & \textbf{89.83} \\

\bottomrule
\end{tabular}

\end{table}

\noindent\textbf{Implementation Details.}  
We use Qwen2.5-VL-7B-Instruct~\cite{bai2025qwen2} as the default frozen vision-language model throughout our experiments, without any model finetuning or adaptation. 
Prompt construction follows a two-step procedure: 
(1) \emph{class-wise fine-grained action based prompts generation}, where an LLM/VLM~\cite{achiam2023gpt,bai2025qwen2} is guided to produce 3--5 action-centric Yes/No questions for each anomaly class, \eg~\enquote{\texttt{Is there any fire or smoke?}} for \enquote{Arson} or \enquote{Exploration} in the UCF-Crime dataset; and 
(2) \emph{prompts compression and summarization
}, where class-specific questions are automatically grouped into semantic clusters via a VLM, with 2--3 generalized guiding questions for each group. 
Detailed prompt templates and the full meta-prompt used for compression are provided in Appendix~C.  In our experiments, we select 6 prompts
for the UCF-Crime dataset and 5 prompts for the XD-violence dataset,
where the generated prompts can be found
in Appendix~D.
Unless otherwise specified, we uniformly extract 128 frames per video segment for inference following conventional practice, and the effect of varying frame numbers is discussed in Appendix~E.  
All experiments are conducted on a single NVIDIA RTX 4090 Ti GPU.
\begin{table}[t]
\renewcommand{\tabcolsep}{3.0mm}
\centering
\small
\caption{AUC performance of VAD methods on XD-Violence.
}
\label{tab:xd-violence}
\begin{tabular}{llc}
\toprule
\textbf{Training Type} & \textbf{Method} & \textbf{AUC\%} \\
\midrule

\multirow{8}{*}{\shortstack{Non-Explainable \\ VAD methods}} 
&Hasan et al.~\cite{hasan2016learning}& 50.32  \\
&RTFM~\cite{tian2021weakly} & 75.89  \\
&CLAP~\cite{al2024collaborative} & 68.60   \\
&FedCoOp~\cite{guo2023promptfl}  & 71.80   \\
&Lu et al.~\cite{lu2013abnormal}& 82.30  \\
&BODS~\cite{wang2019gods} & 83.30  \\
&GODS~\cite{wang2019gods} & 85.99  \\
&RareAnom~\cite{thakare2023rareanom} & 85.62  \\
\midrule
\multirow{7}{*}{\shortstack{Explainable \\ VAD Methods}}

&ZS CLIP~\cite{zanella2024harnessing}& 38.21  \\
&ZS IMAGEBIND-I ~\cite{zanella2024harnessing}&58.81  \\
&ZS IMAGEBIND-V~\cite{zanella2024harnessing}& 55.06  \\
&LLAVA-1.5~\cite{liu2024improved} &79.61\\
&LAVAD~\cite{zanella2024harnessing} &85.36\\
&EventVAD~\cite{shao2025eventvad} &87.51\\
&VERA~\cite{ye2025vera} &88.26\\
&\textbf{ASK-HINT(Ours)} &\textbf{90.31}\\

\bottomrule
\end{tabular}
\end{table}

\subsection{Comparison to State-of-the-Art Methods}
We compare our method with existing approaches on the UCF-Crime and XD-Violence~\cite{sultani2018real,wu2020not} datasets, with results reported in Table~\ref{tab:ucf-crime} and Table~\ref{tab:xd-violence}, respectively.

On UCF-Crime~\cite{sultani2018real}, traditional weakly-supervised methods such as RTFM~\cite{tian2021weakly} and MGFN~\cite{chen2023mgfn} achieve AUC score around 83–86\%, while unsupervised methods (e.g., GCL~\cite{zaheer2022generative}, DYANNet~\cite{thakare2023dyannet}) remain below 85\%. 
On XD-Violence, a similar trend is observed. Classical non-explainable VAD methods such as Hasan et al.~\cite{hasan2016learning} and BODS~\cite{wang2019gods} yield AUC scores ranging from 50\% to 68\%. 
More recently, multimodal LLM-based approaches have pushed the state of the art: fine-tuned Holmes-VAD~\cite{zhang2024holmes} reaches 87.68\%, and HiProbe-VAD (trained in Holmes-VAU)~\cite{cai2025hiprobe} further improves to 88.91\%. 
However, these fine-tuned solutions require substantial computational resources, motivating the exploration of training-free adaptation with frozen VLMs.  
Within the \textit{training-free VLM} category, prior work such as VERA~\cite{ye2025vera}, VADor~\cite{lv2024video}, and HiProbe-VAD report competitive results (85–86\%) in UCF-Crime. 
In XD-violence, recent explainable approaches leveraging vision-language models (e.g., ZS-CLIP~\cite{zanella2024harnessing}, LLaVA~\cite{liu2024improved}, VERA~\cite{ye2025vera}) significantly improve performance, with VERA achieving around 88.26\%. 

Yet, comparing with other existing works, VERA requires additional prompt training with videos, which deviates from a strictly training-free setting. By contrast, our proposed \textit{ASK-HINT} achieves an AUC of \textbf{89.83\%} on UCF-Crime, surpassing all existing training-free methods and even outperforming the best fine-tuned approaches. These results underscore the effectiveness of structured fine-grained prompting in unlocking the anomaly reasoning capabilities of frozen VLMs, while incurring zero additional training cost.  
Overall, these results demonstrate that \textit{ASK-HINT consistently outperforms or matches state-of-the-art methods in the training-free VLM setting}, surpassing fine-tuned solutions. Importantly, our framework achieves this without dataset-specific tuning or computationally expensive optimization, highlighting both its strong generalization ability and practical usability for real-world video anomaly detection.

\subsection{Ablation Study}\label{sec:ablation}

We present a series of ablation studies to systematically evaluate three key factors:
(1) the necessity of prompt selection;
(2) the choice of VLMs;
(3) the number of guiding questions used in $\mathcal{Q}^*$. Unless stated otherwise, all ablation studies are conducted on the UCF-Crime dataset.
\begin{table}[t]
\centering
\renewcommand{\tabcolsep}{5.5mm}
\small
\caption{Comparison between full-prompt baseline and our ASK-HINT framework.
}
\label{tab:full-prompt-comparison}
\begin{tabular}{lcc}
\toprule
\textbf{Method} & \textbf{\#Prompts} & \textbf{AUC\%}  \\
\midrule
Full-Prompt Baseline & 42 & 67.17  \\
ASK-HINT (Ours) & 6 & \textbf{89.83}  \\
\bottomrule

\end{tabular}
\end{table}

\begin{table}[t]
\centering
\caption{Performance with different choice of VLMs,
with and without (baseline) our ASK-HINT prompting strategy.
}
\renewcommand{\tabcolsep}{8.6mm}
\label{tab:backbones}
\small
\begin{tabular}{l|c}
\toprule
\textbf{Model} & \textbf{AUC\%}  \\
\hline
InternVL2.5-8B (baseline)~\cite{chen2024internvl}   & 76.62 \\
InternVL2.5-8B + ASK-HINT   & \textbf{87.42}  \\
InternVideo2.5 (baseline)~\cite{wang2024internvideo2}  & 77.00  \\
InternVideo2.5 + ASK-HINT  & \textbf{89.11}  \\
Qwen2.5-VL-7B (baseline)~\cite{bai2025qwen2}  & 74.50  \\
Qwen2.5-VL-7B + ASK-HINT  & \textbf{89.83}  \\
\bottomrule

\end{tabular}

\end{table}

\noindent\textbf{Directly Inference with $\mathcal{Q}$.} As discussed in Section~\ref{sec:prompt-pool}, a natural baseline is to use $\mathcal{Q}$ directly for video anomaly inference. We report the performance of this baseline in Table~\ref{tab:full-prompt-comparison} as the \enquote{Full-Prompt Baseline}. The results show that using $\mathcal{Q}$ without prompt selection leads to inferior performance, likely due to hallucination effects~\cite{bae2025mash}, highlighting the necessity of prompt compression.

\noindent\textbf{The Choice of VLM.}
To assess the generality of \textsc{ASK-HINT}, we apply it to three frozen vision-language models of varying sizes: InternVL2.5-8B, InternVideo2.5, and Qwen2.5-VL-7B~\cite{chen2024internvl,wang2024internvideo2,bai2025qwen2}. Particularly, we define baselines with abstract prompt, \eg~\enquote{\texttt{Is there any anomaly event?}}.
As shown in Table~\ref{tab:backbones}, our method consistently improves AUC across all evaluated models. For example, compared with the abstract prompt based baselines, \textsc{ASK-HINT} increases AUC by 10.8\% on InternVL2.5-8B, by 12.11\% on InternVideo2.5, and by 15.33\% on Qwen2.5-VL-7B. These results demonstrate the strong generalization capability of our structured prompting approach for VLM-based video anomaly detection.


\noindent\textbf{Effect of Number of Guiding Questions.} 
The number of guiding questions plays a crucial role in shaping the performance of VLMs on the VAD task. 
In our experiments, the compressed prompt set contains 6 guiding questions for the UCF-Crime dataset (Table~\ref{tab:ucf-crime}) and 5 for the XD-Violence dataset (Table~\ref{tab:xd-violence}). 
To further investigate this factor, we edit prompt to compulsoryly vary the number of guiding questions and summarize the results in Table~\ref{tab:questions_ablation}.  
We find that the number of questions strongly affects overall AUC and crime-specific accuracy (\enquote{Crime Acc}, indicating the accuracy of detecting an anomaly video as "Anomaly"), while having relatively little impact on normal video detection. 
Using only 3 questions yields the lowest AUC (78.71\%) and poor crime detection accuracy (61.4\%), indicating insufficient coverage of anomaly patterns. 
Adding more questions (9 or 12) may lead to potential hallucination effects, where longer inputs introduce redundancy, distractive cues, or spurious attention~\cite{bae2025mash}. We select 6 as the final number of guiding questions for the UCF-Crime dataset. This choice is intuitive: since the groups (Step 2 in Section~\ref{sec:semantic-compression}) are already sufficiently separable, a moderate number of summarizing questions (Step 3 in Section~\ref{sec:semantic-compression}) is enough to balance semantic coverage and hallucination mitigation.
\noindent\textbf{VLMs guided $\mathcal{Q}^*$ $\textbf{\textit{vs}}$ Random Selected $\mathcal{Q}^*$.}
To validate the effectiveness of our compression mechanism, we also compare against a random selection of prompts after class-wise prompts construction (Section~\ref{sec:prompt-pool}), and show performance in
Table~\ref{tab:questions_ablation}.
Experimental results with both AUC (\textbf{AUC}) and accuracy (\textbf{Crime Acc}) 
indicate effectiveness of the proposed prompt selection strategy.


\begin{table}[t]
\centering
\renewcommand{\tabcolsep}{2.5mm}
\small
\caption{Ablation study on number of guiding questions. 
We report AUC and crime video detection accuracy for ASK-HINT and random prompt selection. 
}
\label{tab:questions_ablation}
\begin{tabular}{c|c|c|c|c}
\toprule
\multirow{2}{*}{\#\textbf{Questions}} & \multicolumn{2}{c|}{\textbf{ASK-HINT $\mathcal{Q}^*$(\%)}}           & \multicolumn{2}{c}{\textbf{Random $\mathcal{Q}^*$ (\%)}} \\ \cline{2-5} 
                               & \multicolumn{1}{c|}{\textbf{AUC }} & \textbf{Crime Acc} & \multicolumn{1}{c|}{\textbf{AUC}} & \textbf{Crime Acc}                                    \\ 
\hline
3   & 78.71 & 61.43  & 70.10 &42.86 \\
6    & \textbf{89.83} & \textbf{85.00}& 80.83 & 65.00 \\
9   & 87.67 & 80.00  & 81.24 & 67.14\\
12  & 83.36 & 70.71  & 77.88 & 56.43\\
\bottomrule
\end{tabular}

\end{table}

\begin{table}[t]
\centering
\renewcommand{\tabcolsep}{3.0mm}
\caption{
Cross-dataset prompt transfer results (AUC\%) on UCF-Crime and XD-Violence datasets evaluating with AUC Score(\%). 
Rows correspond to test datasets, while columns indicate the prompt sources, comparing with VERA~\cite{ye2025vera}}.
\small
\label{tab:prompt_transfer}
\begin{tabular}{c|cc|c}
\toprule
\multirow{2}{*}{\textbf{Dataset}} 
 & \multicolumn{2}{c|}{\textbf{ASK-HINT (Prompt Source)}} 
 & \multirow{2}{*}{\textbf{VERA}} \\ \cline{2-3}
 & \textbf{UCF-Crime} & \textbf{XD-Violence} & \\ \hline
UCF-Crime   & 89.93 & \textbf{81.86 } & 80.42 \\ 
XD-Violence & \textbf{87.11} & 90.31  & 86.26 \\ 
\bottomrule
\end{tabular}

\end{table}

\subsection{Generalization Analysis}
Most existing prompt optimization approaches (e.g., VERA~\cite{ye2025vera}) are performance-driven, where candidate prompts are selected according to validation accuracy. 
While effective within a given dataset, such prompts are inevitably dataset- or video-specific, raising concerns about their generalization ability. 
In contrast, \textbf{ASK-HINT} derives prompts from action semantics rather than validation scores, yielding a compact set of transferable cues that generalize across unseen classes and datasets.
We evaluate this property under two complementary settings: cross-dataset transfer and cross-class transfer. 

\noindent\textbf{Cross-Dataset Transfer.} 
We first study whether prompts constructed from one dataset can be applied to another dataset, and show performance of AUC score in 
Table~\ref{tab:prompt_transfer}. 
Table~\ref{tab:prompt_transfer} reports results on UCF-Crime and XD-Violence datasets, where prompts are constructed from one dataset and applied to the other. 
For fairness, we also evaluate VERA~\cite{ye2025vera} in the cross-dataset setting, where its performance on UCF-Crime is obtained using prompts derived from XD-Violence, and vice versa.
The results show that \textbf{ASK-HINT} consistently outperforms VERA when transferring across datasets. 
On UCF-Crime, ASK-HINT achieves 81.86\% AUC with prompts from XD-Violence, compared to 80.42\% with VERA. 
On XD-Violence, ASK-HINT achieves 87.11\% AUC with prompts from UCF-Crime, again surpassing VERA (86.26\%). 
These results confirm that prompts derived from fine-grained action semantics generalize better than those obtained via validation-driven optimization. 
Notably, ASK-HINT achieves higher transferability without relying on validation accuracy, highlighting its superiority
in training-free settings.

\noindent\textbf{Cross-Class Transfer.} 
We further investigate whether prompts designed for a subset of anomaly classes can generalize to other classes within the same dataset. 
To construct the subset of seen categories, we leverage the semantic clustering results (Figure~\ref{fig:heatmap-prompts}) and randomly select 1–2 representative classes from each cluster. 
The intuition is that if different anomaly types share common fine-grained action primitives, then prompts derived from representative classes should be able to transfer to the remaining unseen classes. 
Concretely, for the UCF-Crime dataset, we define the seen set $V_{seen}$ as consisting of \textit{Arson, Road Accident, Explosion, Robbery, Arrest, Assault, Stealing}, 
and the unseen set $V_{unseen} = V_{test} \setminus V_{seen}$. 
We then generate a compact prompt set $\mathcal{Q}^*_{seen}$ from $V_{seen}$ and evaluate it on both subsets, while also reporting overall performance on the entire $V_{test}$ for completeness. We also include the baseline performance with abstract prompt for clear comparison.

\begin{table}[t]
\centering
\renewcommand{\tabcolsep}{4.0mm}
\caption{Cross-class transfer performance using prompts generated from a subset of classes. 
Results are reported for all test videos, seen classes, and unseen classes. }
\small
\label{tab:cross_class_results}
\begin{tabular}{lcc}
\toprule
\textbf{Setting} & \textbf{ASK-HINT} & \textbf{Abstract} \\
\midrule
\textbf{All Test (AUC\%) }       & \textbf{84.38} & 80.28 \\
\textbf{Seen Classes (Acc\%)}  & \textbf{74.60} & 44.44 \\
\textbf{Unseen Classes (Acc\%)}& \textbf{61.03} & 31.16 \\
\bottomrule
\end{tabular}
\end{table}

As shown in Table~\ref{tab:cross_class_results}, \textbf{ASK-HINT} achieves an AUC of 84.38\% across all test videos, outperforming abstract prompting (78.00\%). 
For seen classes, ASK-HINT yields an accuracy of 74.60\% over crime detection, compared to only 44.44\% with abstract prompts. 
More importantly, on unseen classes, ASK-HINT still achieves 61.03\% accuracy, nearly doubling the abstract baseline (31.16\%). 
These results confirm that the fine-grained action semantics captured by ASK-HINT encode transferable primitives (e.g., \emph{physical confrontation}) that recur across anomaly categories. 



\subsection{Qualitative Results and Case Studies}\label{Sec:quali}

Another key advantage of our method lies in its strong interpretability (see Figure~\ref{fig:prompt_template}). We further analysis
interpretability of our solution with case studies.
We organize the case studies into two parts: (1) representative examples from seen anomaly classes, and (2) an unseen class case study that demonstrates cross-class generalization.
\begin{figure}[t]
    \centering
    \includegraphics[width=0.95\linewidth]{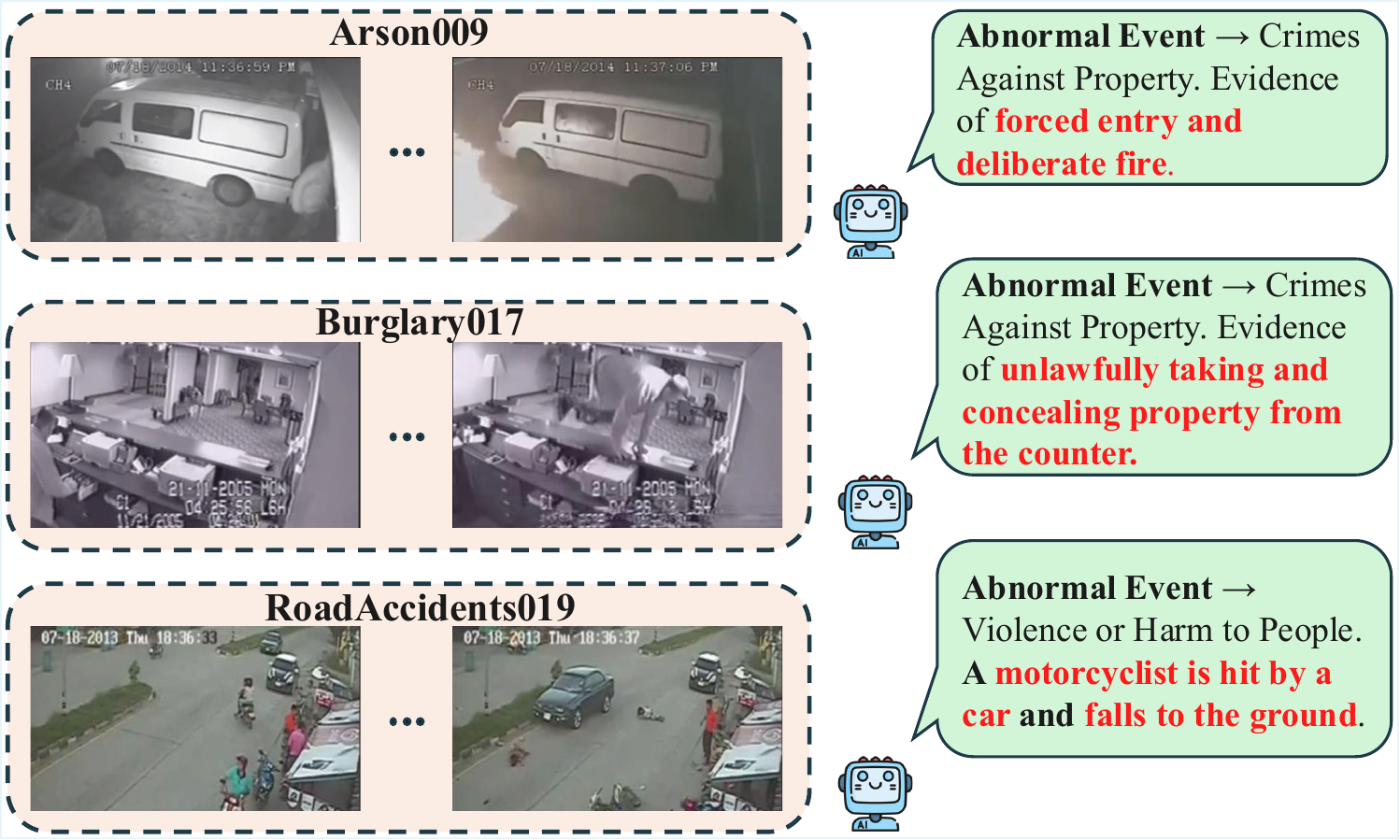}
    \caption{Qualitative case studies on UCF-Crime dataset, where ASK-HINT not only detects abnormal events but also provides reasoning traces aligned with fine-grained action semantics.}
    
    \label{fig:qualitative}
\end{figure}
\begin{figure}[t]
    \centering
    \includegraphics[width=0.95\linewidth]{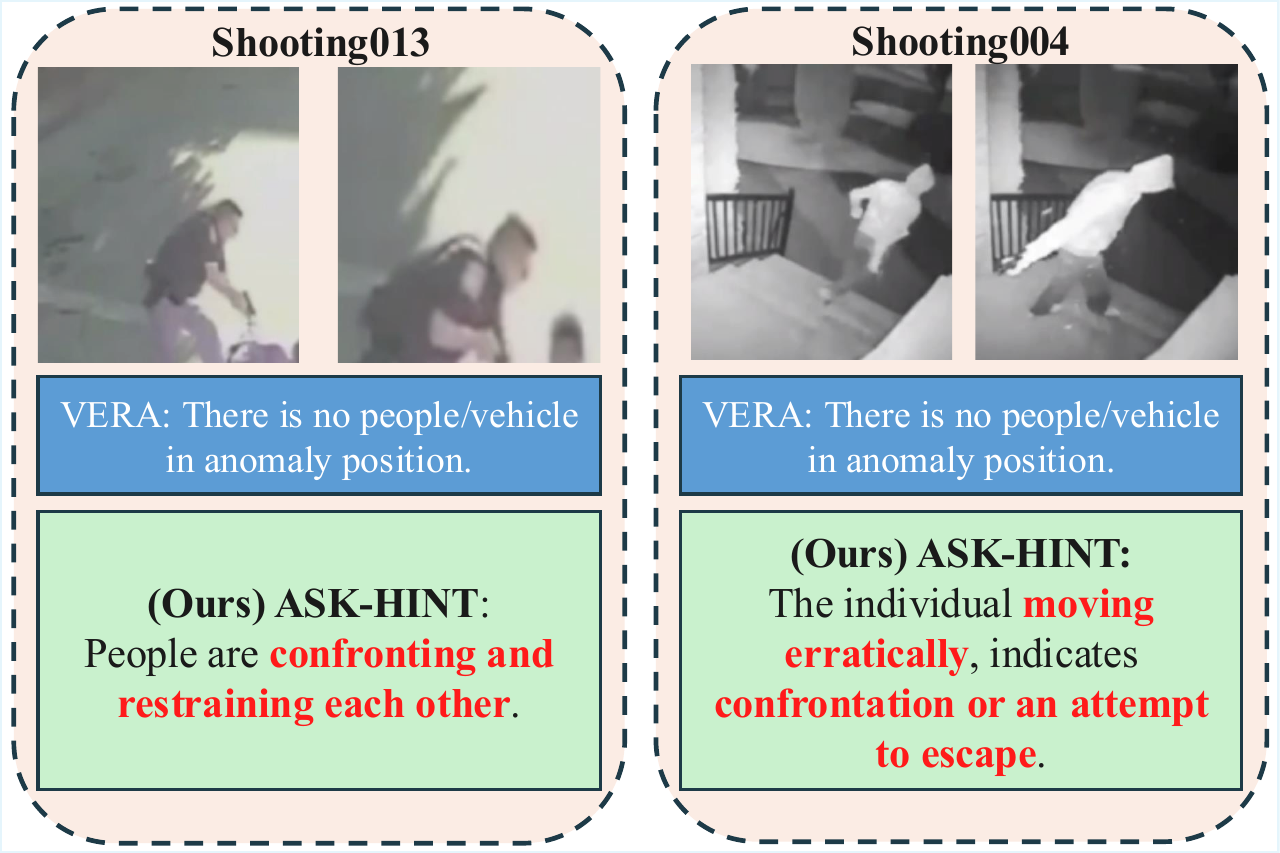} 
    \caption{Case Study on Unseen Class shooting.
    Although “Shooting” are not included in either VERA or ASK-HINT prompts, the results show clear differences in generalization. 
    where
    \textit{VERA} (middle) fails to detect any anomaly. 
    and \textit{ASK-HINT} (bottom) successfully identifies fine-grained anomaly cues. 
    }
    \label{fig:qual_shooting}
\end{figure}

\noindent\textbf{Representative Cases on UCF-Crime.}
Figure~\ref{fig:qualitative} presents three representative examples. 
In \textit{Arson009}, ASK-HINT highlights ``forced entry and deliberate fire'', providing a transparent explanation for the abnormal classification. 
In \textit{Burglary017}, it identifies ``unlawfully taking and concealing property'', aligning with human interpretation. 
In \textit{RoadAccidents019}, the model explains that ``a motorcyclist is hit by a car and falls to the ground''. 
These structured reasoning traces demonstrate how ASK-HINT transforms VLM outputs
into human-auditable explanations.

\noindent\textbf{Unseen Class Generalization.}
One of the most compelling aspects of ASK-HINT is its ability to generalize beyond explicitly defined categories. 
We analyze the ``Shooting'' class, whose related prompts are \emph{not included} in the prompt sets of either VERA or ASK-HINT. 
As shown in Figure~\ref{fig:qual_shooting}, VERA~\cite{ye2025vera} fails to detect any anomaly, outputting vague statements such as ``no people/vehicle in anomaly position''. 
In contrast, ASK-HINT captures transferable action primitives, such as ``confrontation'' and ``restraining'', which indirectly characterize the shooting context. 
This case highlights that ASK-HINT does not rely on memorizing prompts but instead reuses fundamental action cues across anomaly categories, enabling zero-shot generalization to unseen anomalies. 


\section{Conclusion}

We presented \textbf{ASK-HINT}, a structured prompting framework for video anomaly detection with frozen VLMs. By introducing fine-grained, action-centric questions organized into semantic groups, ASK-HINT enables interpretable reasoning and outperforms existing training-free and even fine-tuned baselines on UCF-Crime and XD-Violence. Our experiments show that a compact set of carefully designed prompts strikes the best balance between coverage, stability, and accuracy.  
Beyond accuracy, ASK-HINT offers transparent explanation traces and strong generalization, including cross-dataset transfer and detection of unseen anomalies. These results highlight structured prompting as a simple yet effective alternative to fine-tuning, making it practical for open-world anomaly detection. Future work will explore extending this framework to broader video understanding tasks and dynamic, context-aware prompting.

\paragraph{Limitations and Future Work.}
Despite its effectiveness, 
ASK-HINT has several limitations. 
First, it relies on a static prompt set derived offline, which may not fully capture
novel anomalies in dynamic environments. 
Second, our framework ignore
temporal modeling, showing limitations
to reason over evolving events. 
Future work will explore \textit{dynamic, context-aware prompting} that adapts to video content.
Incorporating temporal reasoning, multimodal cues, and human refinement offers promising directions, as does evaluating the framework in open-world settings.



{
    \small
    \bibliographystyle{ieeenat_fullname}
    \bibliography{main}
}

\clearpage

\appendix

\section{Class-Wise Multi-Granularity of Prompts}\label{appen:Class_Granular_Prompt}

In this section, we aim in using prompts from three-level of granularity to verify our hypothesis where the fine-grained actions can improve frozen VLMs in VAD task. Here we provide the following example. All the actions are generated by GPT~\cite{achiam2023gpt} Here is the example:

\begin{tcolorbox}[title=Abstract Prompt]
Please analyze the following video step-by-step and determine whether it contains abnormal behavior.  
Answer Yes or No with a short description on the video.
\end{tcolorbox}

\begin{tcolorbox}[title=Group-Level Prompt]
Considering the following group knowledge:
- Violence or Harm to People  

- Crimes Against Property  

- Public Safety Incidents  

Based on the understanding, does this video depict a \textbf{Stealing} event?  
Answer Yes or No, and explain briefly.
\end{tcolorbox}

\begin{tcolorbox}[title=Fine-Grained Prompt (\textit{Stealing})]
1. Are there people taking items without permission? \\
2. Do individuals appear to be carrying or moving objects away from a specific location? \\
3. Is there a visible struggle or resistance between individuals? \\
4. Are there signs of hiding or concealing objects? \\
5. Do people seem to be looking around suspiciously while handling items?
\end{tcolorbox}

\section{Na\"{i}ve Basline in Using all Prompts}\label{appen:naive_method}

\paragraph{Motivation.} One natural baseline is to apply the entire class-wise prompt pool $\mathcal{Q}$—comprising all fine-grained action prompts for every anomaly class—to every test video during inference. This strategy ensures that the model has access to all potentially relevant reasoning cues, regardless of the specific anomaly type in the video.

\paragraph{Implementation.} During inference, each video segment is prompted with the full list of questions $\mathcal{Q}$, embedded into a single prompt template. The VLM is required to answer all questions and make a final binary decision (normal/abnormal) based on the collective reasoning.

\paragraph{Details of the prompt.} This is the code for 

\begin{tcolorbox}[title= Entire Pool to for Prompt Generation]
You are analyzing ONE surveillance video.  

For EACH of the 14 classes below, answer THREE class-specific diagnostic questions with ``Yes" or ``No". Then give a short reason ($<$12 words) and a confidence score in [0,1] for that class's overall decision (``answer": Yes/No).  
Finally, output the final answer on whether there is an anomaly event. Answering ``yes" or ``no".  

\textbf{CLASSES AND QUESTIONS}  

1) Robbery  

Q1: Is there direct confrontation between aggressor and victim?  

Q2: Is force/threat/intimidation involved (\eg, weapon, restraint)?  

Q3: Is property taken during or right after the confrontation? 

...

13) Vandalism

Q1: Is property deliberately damaged (smash/graffiti/scratch/break)?

Q2: Are objects/vehicles/buildings targeted (not people)?

Q3: Is the damage intentional, not accidental?

14) Normal Event

Q1: Are people calm without violence/ theft/ accidents/ hazards?

Q2: Is property intact with no damage/ tampering/ fire?

Q3: Are movements typical daily life (walking/ shopping/ waiting)?

\textbf{OUTPUT FORMAT (STRICT):}  
Return ONLY raw JSON (no markdown, no extra text) with this schema:  
\{  
  ``final\_label": ``Yes" or ``no"  
\}  

\textbf{RULES:}  

- Evaluate ALL 14 classes and ALL their questions.  

- ``answer" is your overall Yes/No for that class, consistent with Q1–Q3.  

- Confidence reflects visual evidence strength for that class.  

- Keep reasons short ($<$12 words).  

- Output valid JSON only.  
\end{tcolorbox}

\section{Meta-Prompt for Prompt Generation and Compression}\label{appen:meta_prompt}
In this section, we aim in providing the prompt for LVLM to generate class-wise fine-grained actions and summarize them into a compact set of prompts.

\begin{tcolorbox}[title=Prompt for Shared Action Compression]
You are an \textbf{expert in video anomaly detection using Vision-Language Models}.  
Your task has two steps:

\textbf{Step 1: Generate class-specific guiding questions}  
For each anomaly class in the list, generate \textbf{3--5 short, Yes/No guiding questions}.  
\begin{itemize}
    \item The questions must be \textbf{action-centric} and \textbf{context-aware} (\eg, ``Do you see people fighting?'').  
    \item They should help a model distinguish the target anomaly class from others and from normal events.  
    \item Output each class with its list of questions.  
\end{itemize}

\textbf{Anomaly Classes:}  
\begin{itemize}
    \item Abuse  
    \item Car Accident  
    \item ... 
    \item Riot  
\end{itemize}

\textbf{Step 2: Summarize and Conclude}  
Your task is to \textbf{summarize and group} these guiding questions into a compact set.  

\textbf{Steps:}  
\begin{enumerate}
    \item Read all the class-specific guiding questions.  
    \item Cluster them into \textbf{major groups} based on similar actions or themes.  
    \item For each group, summarize the questions and generate \textbf{2--3 generalized guiding questions} in Yes/No format, capturing the common patterns from the original class prompts.  
    \item Avoid vague words like ``abnormal'' — use \textbf{action- or object-specific terms} (\eg, ``fighting,'' ``stealing,'' ``breaking,'' ``explosion'').  
    \item Provide a \textbf{compact final set} of grouped guiding questions.  
\end{enumerate}

\textbf{Output Format:}  

Grouped Guiding Questions:  

\textbf{Group 1: [Group Name]}  
1. ...  
2. ...  
3. ...  

\textbf{Group 2: [Group Name]}  
1. ...  
2. ...  
3. ...  

\textbf{Group 3: [Group Name]}  
1. ...  
2. ...  
3. ...  

\textbf{Summary:} [One sentence explaining what these grouped guiding questions aim to achieve]  
\end{tcolorbox}

\paragraph{Limitations.} Two key issues with this strategy:
\begin{itemize}
    \item \textbf{Prompt length exceeds model context window}: For models like Qwen2-VL and InternVL2, the number of tokens required to encode all prompts and video context often exceeds the maximum input length, leading to truncation or memory errors.
    \item \textbf{Increased hallucination and reduced focus}: When too many irrelevant prompts are included (\eg, fire-related questions on a theft video), the model often generates noisy or inconsistent responses, degrading accuracy.
\end{itemize}

\paragraph{Empirical Results.} Table in main paper compares ASK-HINT against the full-prompt baseline. Despite using fewer prompts, ASK-HINT achieves higher performance due to semantic compression and improved alignment.

\begin{figure*}[t]
    \centering
    \includegraphics[width=\textwidth]{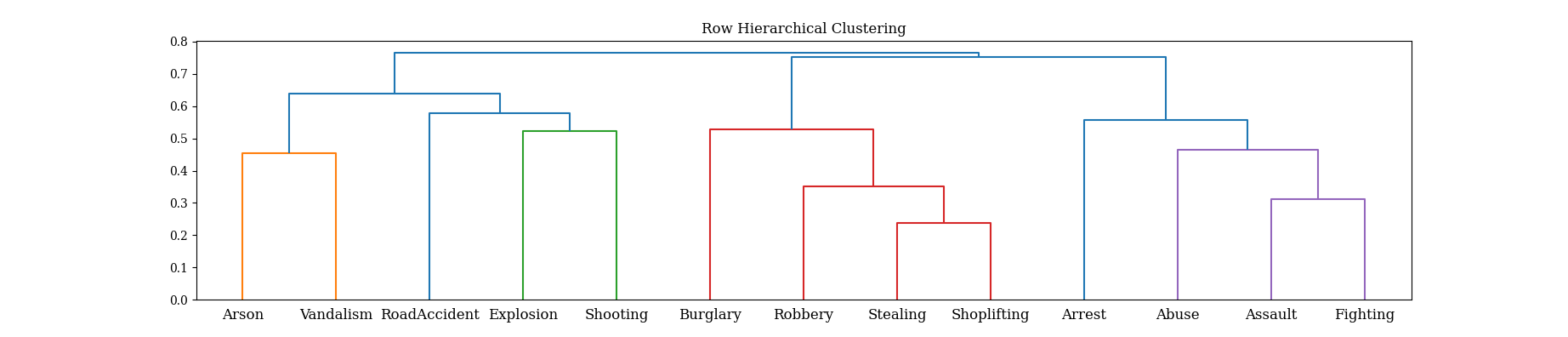}
    \caption{Hierarchical clustering of UCF-Crime categories based on fine-grained action semantics. 
    The dendrogram reveals meaningful groupings, such as \textit{Assault}, \textit{Fighting}, and \textit{Abuse} 
    (all involving direct human confrontation), or \textit{Arson} and \textit{Explosion} (both involving fire-related actions). 
    These connections motivate our structured prompting framework, which leverages shared action primitives across anomaly classes 
    to construct a compact and generalizable prompt set.}
    \label{fig:ucfcrime_dendrogram}
\end{figure*}
\section{Prompt for XD-Violence}

\begin{tcolorbox}[title=\textbf{ASK-HINT Prompt for XD-Violence}]
\small
\textbf{Instruction:} You are analyzing one surveillance or online video.  

\textbf{Task 1:} Decide if the video is \textit{Normal} or \textit{Abnormal}.  

\textbf{Task 2:} If \textit{Abnormal}, consider the following guiding questions to identify violent or hazardous events:

\begin{itemize}
    \item Q1: Do you see people engaging in physical conflict such as hitting, kicking, or grappling?  
    \item Q2: Is someone being restrained, abused, or violently controlled by others?  
    \item Q3: Do you observe firearms, gunfire, or threats with visible weapons?  
    \item Q4: Are there signs of explosions, fire outbreaks, or large-scale destruction?  
    \item Q5: Do you see vehicles crashing, losing control, or hitting people or structures?  
\end{itemize}

\textbf{Answer format:}  
\begin{itemize}
    \item ``Normal Event. [short reason]''  
    \item ``Abnormal Event. [short reason referencing Q1--Q5]''  
\end{itemize}
\end{tcolorbox}

\section{Effect of Number of Frames}
To study the influence of temporal granularity, we vary the maximum number of sampled frames for each video segment and report the performance in terms of AUC, correct prediction rate on abnormal (crime) videos, and correct prediction rate on normal videos. The results are summarized in Table~\ref{tab:frames_ablation}. 

We observe that the AUC remains relatively stable across different settings, ranging from 0.888 to 0.898, seeing result in Table~\ref{tab:frames_ablation}. Interestingly, both small (\eg, 8 frames) and large (\eg, 256 frames) sampling configurations achieve competitive performance, while intermediate settings (\eg, 32 frames) slightly degrade the accuracy. For crime videos, the correct detection rate is consistently around 0.84--0.85, suggesting that the model is robust in recognizing abnormal events regardless of the number of frames. For normal videos, increasing the number of frames provides a marginal benefit, improving the correct classification rate.

Moreover, the number of frames reported here refers to the \emph{maximum} sampled frames. If a video is shorter than the target length, we uniformly extract frames at $fps=1$ until all available frames are used. This strategy ensures consistency across videos of different durations while avoiding artificial duplication or bias.

\begin{table}[t]
\centering
\caption{Ablation study on the number of sampled frames. 
We report frame-level AUC, correct prediction rate on crime videos, 
and correct prediction rate on normal videos.}
\label{tab:frames_ablation}
\begin{tabular}{c|c|c}
\hline
\#Frames & AUC (\%)& Crime Correct (\%)\\
\hline
8   & 89.14 & 84.29 \\
16  &89.47 & 84.29 \\
32  &88.79 & 83.57 \\
64  & 89.14 & 84.29 \\
128 & 89.83 & 85.00 \\
256 & 89.83 & 85.00 \\
\hline
\end{tabular}
\end{table}

\section{Hierarchical Connection for UCF-Crime}
To better understand the semantic relationships among anomaly categories in UCF-Crime, 
we conduct a hierarchical clustering analysis based on the similarity of their fine-grained action prompts. 
As shown in Fig.~\ref{fig:ucfcrime_dendrogram}, the dendrogram reveals several meaningful groupings. 
For example, \textit{Assault}, \textit{Fighting}, and \textit{Abuse} are closely clustered, reflecting their shared reliance on physical confrontation cues. 
Similarly, \textit{Arson} and \textit{Explosion} are linked by fire-related actions, while \textit{Robbery}, \textit{Stealing}, and \textit{Shoplifting} are grouped together through theft-related behaviors.

This hierarchical structure highlights two important insights. 
First, anomaly categories are not independent but often share underlying action primitives, 
suggesting that prompts can be compressed into a smaller representative set without losing semantic coverage. 
Second, these shared connections provide stronger interpretability: by tracing model predictions back to clusters of fine-grained actions, 
we can explain why different anomaly classes exhibit related reasoning patterns. 
Together, this motivates our design of ASK-HINT, which leverages hierarchical connections to construct a compact and generalizable prompt set $\mathcal{Q}^*$.

\end{document}